\documentclass[]{fairmeta}
\title{PhysCreator: Creating Physically-Attributed 3D Assets at Will}

\author[1]{Shiyan Zheng}
\author[2]{Ming Li}
\author[3]{Lei Han}
\author[1*]{Xiangyu Xu}

\affiliation{$^1$Xi'an Jiaotong University}
\affiliation{$^2$The Chinese University of Hong Kong, Shenzhen}
\affiliation{$^3$Discover Robotics}

\contribution[*]{Corresponding author}

\abstract{
Significant advancements in 3D asset creation have enabled high-fidelity 3D generation from images. However, most existing representations remain inherently static, lacking the ability to model physically plausible behaviors under dynamic interactions. We present \textbf{PhysCreator}, a unified framework that directly generates simulation-ready 3D representations with spatially varying physical properties from images. Unlike prior approaches that treat physical attributes as simple post-hoc predictions or rely on expensive per-instance optimization, we reformulate physical property estimation within a unified 3D representation learning framework. We introduce a voxel-aligned latent space by lifting image observations into a spatially structured 3D representation, in which geometry, appearance, and physical properties are jointly modeled and inferred in a feed-forward manner. To enable scalable learning without densely annotated physical ground truth, we leverage vision-language distillation to transfer material-aware semantic priors from powerful vision-language models into spatially grounded voxel-level supervision. Experiments demonstrate that our method achieves superior visual quality, physical plausibility, and generalization compared to prior approaches.  
}

\usepackage{hyperref}
\usepackage{url}
\usepackage[T1]{fontenc}

\usepackage{booktabs}
\usepackage{amsfonts}
\usepackage{nicefrac}
\usepackage{microtype}

\usepackage{epsfig, xspace, enumitem}

\usepackage{
  graphicx,
  amsmath,
  amssymb,
  caption,
  subcaption,
  multirow,
  overpic,
  textpos,
  pifont,
  adjustbox
}

\usepackage{xcolor}
\usepackage[utf8x]{inputenc}

\makeatletter
\@namedef{ver@everyshi.sty}{}
\makeatother

\usepackage{pgf, tikz, pgfplots}
\usepackage{makecell}
\usepackage{pgf-pie}
\usepackage{float}
\usepackage{wrapfig}
\usepackage{colortbl}

\usepackage{newfloat}
\usepackage{listings}
\usepackage{siunitx}
\usepackage{amsthm}

\usepackage{comment}
\usepackage{threeparttable}

\DeclareUnicodeCharacter{3010}{[}
\DeclareUnicodeCharacter{3011}{]}

\theoremstyle{definition}

\theoremstyle{remark}

\setlist[enumerate]{
  itemsep=-0.5mm,
  partopsep=0pt
}

\pgfplotsset{compat=1.18}

\begin{document}

\maketitle

\begin{figure*}[t]
\includegraphics[width=\linewidth]{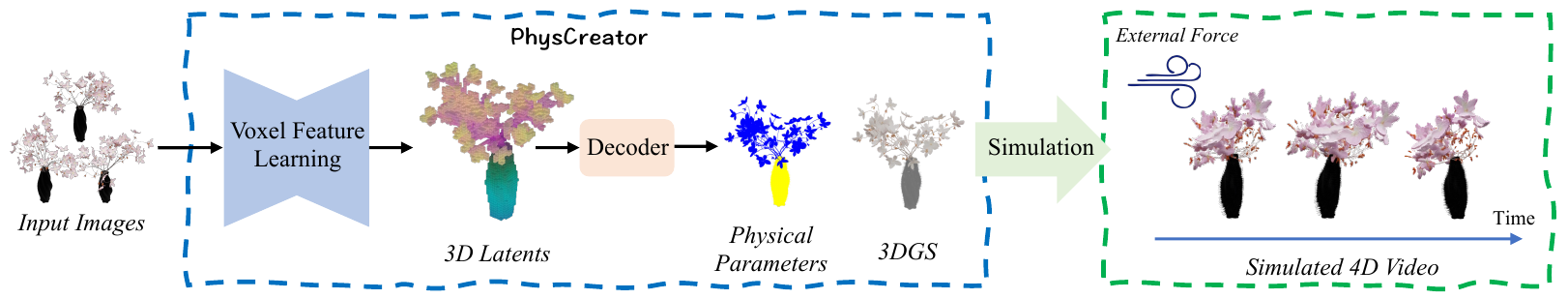}
\caption{
We present PhysCreator, a physics-aware 3D generative model that
produces simulation-ready 3D Gaussians with spatially-varying, semantically-consistent physical properties directly from images.
Our method enables the joint prediction of geometry and physically
meaningful attributes, such as material type, stiffness, and density,
with strong generalization across diverse objects.
The generated 3D Gaussians can be seamlessly integrated into simulations, yielding realistic and dynamic physical behaviors.
}
\label{fig:teaser}
\end{figure*}

\section{Introduction}
Recent advances in 3D generation and reconstruction have enabled the creation of high-fidelity visual assets from images, significantly improving the realism of virtual environments. Methods based on neural rendering~\citep{mildenhall2021nerf,poole2022dreamfusion} and 3D Gaussian representations~\citep{kerbl20233d,li2025asymgs} can produce photorealistic geometry and appearance, supporting applications in graphics, robotics, and simulation. 
However, these representations remain inherently static: while they capture how objects look, they do not model how they behave under physical interactions. Enabling simulation-ready 3D assets therefore requires not only geometry and appearance, but also physically meaningful properties such as stiffness and density.

Inferring such physical properties from images is challenging, as quantities such as Young’s modulus and density are not directly observable.
Existing approaches either rely on per-instance optimization~\citep{zhang2024physdreamer, huang2025dreamphysics,lin2025omniphysgs} or adopt modular pipelines for separate reconstruction and semantic reasoning~\citep{lin2025phys4dgen}.
A key limitation underlying these approaches is that geometry reconstruction and physical property estimation are typically performed in separate stages, rather than being jointly modeled within a unified representation.
Consequently, they often suffer from inefficient inference, limited generalization, and inconsistent geometry–physics coupling.
Recent work such as PhysGM~\citep{lv2025physgm} takes a step toward addressing this issue by jointly predicting 3D geometry and physical properties in a feedforward manner, which improves geometric-physical consistency and enables immediate simulation and 4D rendering.
However, it operates primarily in a 2D feature space, mapping pixels to 3D primitives and using global features for physical prediction. This formulation lacks explicit 3D spatial structure, making it difficult to model spatially varying physical properties and leading to oversimplified, globally uniform materials.

In this work, we propose \textbf{PhysCreator}, a unified framework that jointly predicts geometry and spatially varying physical properties from images, enabling efficient and generalizable generation of simulation-ready 3D assets, as illustrated in Fig.~\ref{fig:teaser}. 
Our key idea is to lift visual observations into a 3D feature representation, which provides explicit spatial support for both geometry and physics.
From this representation, we jointly predict 3D Gaussian primitives for geometry and appearance, along with per-voxel physical attributes such as Young’s modulus, density, and Poisson’s ratio. This design enables fine-grained, spatially-consistent material modeling.
Modeling spatially varying physical properties introduces additional challenges for training, as it requires fine-grained physical supervision signals, which are difficult to obtain at scale.
To address this challenge, we introduce a vision-language distillation strategy that transfers material-aware knowledge from pretrained vision-language models (VLMs) into voxel-level physical supervision.
Specifically, the VLM identifies semantic object parts and their associated physical attributes, which are subsequently lifted into the shared 3D space to supervise the physical property decoder.
This design enables PhysCreator to learn generalizable mappings from visual appearance to fine-grained, spatially consistent material attributes without requiring densely annotated physical ground truth. Moreover, PhysCreator supports optional semantic conditioning, allowing high-level material descriptions to guide the predicted physical-property values without manual numerical tuning.

Our approach offers several strengths: 
1)~joint prediction of geometry and spatially varying physical properties, enabling high-fidelity, controllable, and simulation-ready material modeling; 
2)~scalable supervision via efficient 3D representations and vision-language models, enabling fine-grained physical learning without dense manual physical annotations; and
3)~generalizable feed-forward inference for both multi-view reconstruction and single-view generation.

We demonstrate that PhysCreator produces high-quality 3D assets with physically plausible behaviors and enables realistic dynamic simulation across a wide range of scenes. Our results highlight the potential of combining 3D modeling, physical reasoning, and vision-languager priors to bridge the gap between visual reconstruction and physical understanding.

\section{Related Work}

\subsection{3D Reconstruction and Generation}

\paragraph{3D reconstruction.} Reconstructing 3D assets from multi-view images is a fundamental problem in computer vision~\citep{hartley2003multiple, oliensis2000critique}, traditionally addressed by Structure-from-Motion (SfM)~\citep{agarwal2011building, frahm2010building, wu2013towards, schonberger2016structure} and Multi-View Stereo pipelines~\citep{furukawa2015multi, fu2022geo, gu2020cascade}. These approaches estimate camera parameters and recover scene geometry through multi-stage optimization, achieving high-fidelity reconstruction under well-constrained settings. 
More recently, learning-based methods directly predict 3D structures from images, including point clouds~\citep{lionar2023nu,brachmann2024scene, wang2024vggsfm, wang2024dust3r, wang2025vggt}, depth maps~\citep{Gan_2018_ECCV,depthanything,he2026depthclip3d}, meshes~\citep{xu20203d,xu2023smpler,liu2023towards}, neural fields~\citep{mildenhall2021nerf,chen2022geometry,liang2023infnerf}, and 3D Gaussian Splatting~\citep{kerbl20233d,li2025asymgs}, significantly simplifying the reconstruction pipeline while maintaining strong geometric accuracy.

\paragraph{3D generation.} Complementary to reconstruction, 3D generation has rapidly advanced with the development of modern generative models. Diffusion-based approaches have become the dominant paradigm, enabling the synthesis of diverse 3D representations~\citep{zhao2023michelangelo, chen2023single, luo2021diffusion, hu2024neural,li2024instant3d, he2024gvgen, wu2025direct3d}. In addition, autoregressive models inspired by large language models have been explored for structured 3D generation~\citep{radford2019language, chen2025mar}. Recent works further aim to unify different latent representations into a shared framework, improving both generation quality and flexibility~\citep{xiang2025structured}. Despite these advances, existing methods primarily focus on geometry and appearance, while largely overlooking the physical properties of generated 3D assets.

\subsection{Physics-Based 3D Simulation}

Recent works aim to combine physics simulation with neural 3D representations for generating simulation-ready dynamic assets~\citep{xie2024physgaussian, zhang2024physdreamer}. 
Most existing approaches follow a multi-stage pipeline, where geometry is first reconstructed and physical properties are then estimated for downstream simulation.

Existing approaches either rely on per-instance optimization~\citep{zhang2024physdreamer, huang2025dreamphysics, lin2025omniphysgs} or adopt modular pipelines for separate reconstruction and semantic reasoning~\citep{lin2025phys4dgen, le2025pixie}. 
While optimization-based methods are computationally expensive and difficult to generalize, modular approaches often suffer from weak geometry--physics coupling and limited physical consistency.
Recent work such as PhysGM~\citep{lv2025physgm} takes a step toward joint modeling by directly predicting geometry and physical properties in a feed-forward manner. 
However, it primarily operates in a 2D feature space and predicts physical properties from global features, lacking explicit 3D spatial structure for modeling heterogeneous materials. 
As a result, these feedforward methods often produce spatially coarse or globally uniform physical properties, limiting their ability to capture fine-grained material variations and realistic physical behaviors.

In contrast, we model geometry and physical attributes within a unified voxel-aligned latent representation, enabling spatially varying physical properties and simulation-ready 3D generation.

\begin{figure*}[t]
  \centering
  \includegraphics[width=\textwidth]{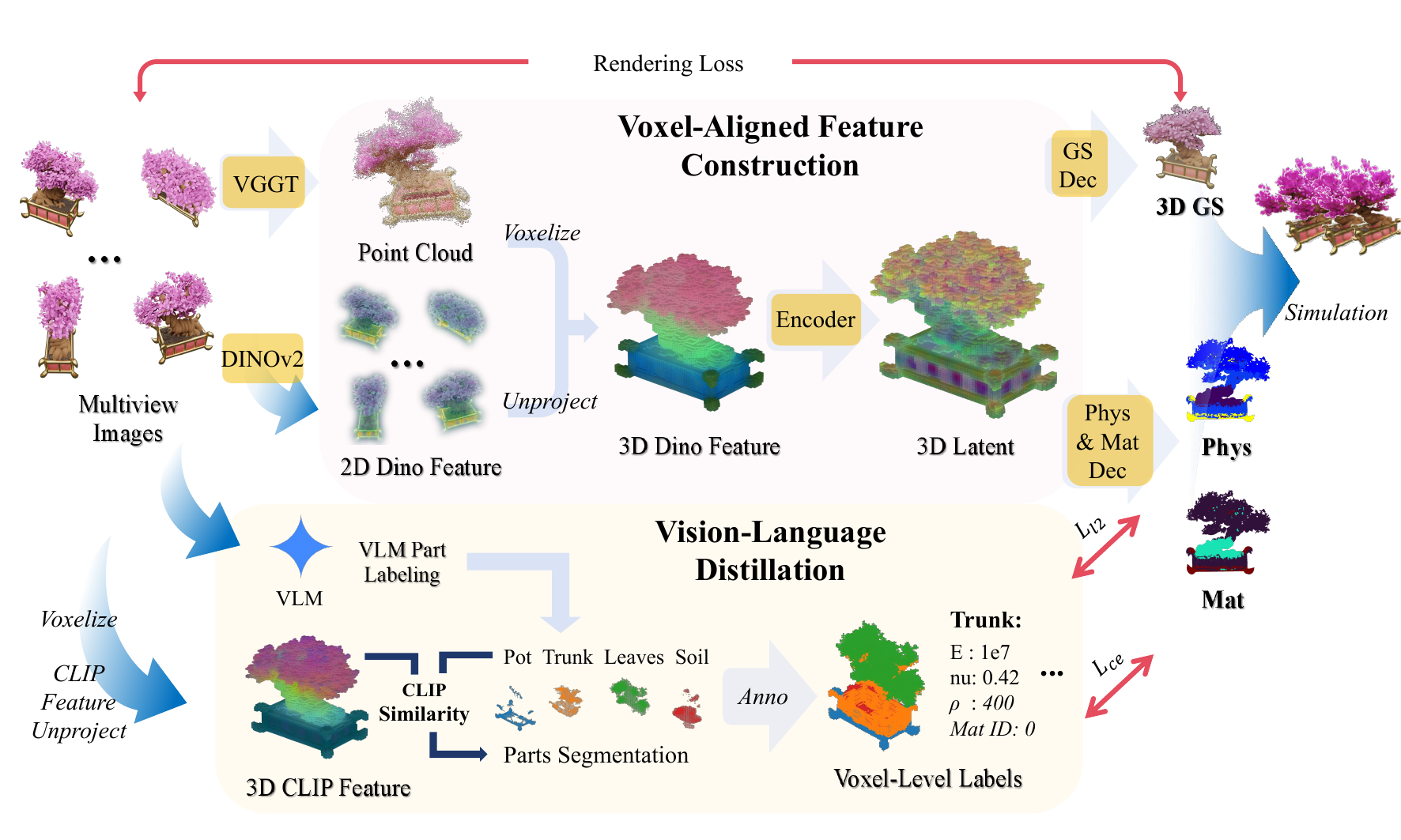}
  \caption{
{Overview of PhysCreator}. \textbf{Top}: Given observations, our framework first constructs a voxel-aligned latent representation (the 3D latent). Built upon this shared representation, we decode simulation-ready 3D Gaussian primitives together with voxel-wise physical attributes, including Young’s modulus, density, Poisson’s ratio, and material category. 
\textbf{Bottom}: To provide scalable supervision, vision-language distillation generates part-level semantic and physical annotations, which are lifted into voxel-level labels for training the physical-material decoder. The predicted geometry and physical properties are then passed to an MPM simulator to evaluate and visualize dynamic behavior.
}
  \label{fig:method-overview}
\end{figure*}

\section{Method}
Given one or more input images, our goal is to reconstruct a simulation-ready 3D representation that jointly models geometry, appearance, and spatially varying physical properties. 
To this end, we propose PhysCreator, a unified framework that combines a voxel-aligned latent representation (Sec.~\ref{sec:unified geo-phys learning}) with scalable
vision-language physical supervision~(Sec.~\ref{Vision-Language Distillation}).

As illustrated in Fig.~\ref{fig:method-overview}, our method consists of two key components.
First, we construct a voxel-aligned latent representation from multi-view observations and learn to generate the same representation from a single image using conditional flow matching. 
This shared representation is then jointly decoded into 3D Gaussian primitives and spatially varying physical properties.
Second, we distill material-aware knowledge from pretrained vision-language models into spatially grounded, voxel-level supervision for learning the physical properties.
Together, these components enable PhysCreator to generate geometrically aligned and spatially varying physical attributes through efficient feed-forward inference. The physical decoder can further use optional semantic conditioning to guide the predicted physical-property values without manual numerical tuning.

\subsection{Unified Geometry-Physics Learning}\label{sec:unified geo-phys learning}

Directly predicting both geometry and physical properties from images is inherently challenging, particularly for spatially varying materials.
To the best of our knowledge, PhysGM~\citep{lv2025physgm} is among the few works that attempt to address this problem.
%
However, it operates primarily in a 2D feature space, where geometry is inferred from pixel-wise predictions while physical properties are predicted from global image features~\citep{lv2025physgm}. 
This formulation lacks explicit 3D spatial structure, making it difficult to associate physical properties with specific regions in 3D. 
As a result, it relies on simplified assumptions such as globally uniform materials and struggles to capture fine-grained details.

To address this limitation, we introduce a voxel-based representation that lifts visual observations into a 3D feature space. 
This design provides explicit spatial support and locality, enabling both geometry and physical properties to be predicted in a shared 3D domain.
By grounding both predictions in the same spatial structure, our method naturally supports spatially varying physical properties and ensures consistency between geometry and physics. 
In addition, it facilitates the reconstruction of detailed and high-fidelity geometry, providing a strong foundation for both accurate rendering and physically consistent simulation (as illustrated in Fig.~\ref{fig:vs_physGM}).


\subsubsection{Voxel-Based Feature Construction}
\label{Voxel-Aligned Feature Construction}

Given multi-view images $\{\mathbf{I}_i\}_{i=1}^N$, we reconstruct a consistent 3D geometry using a pretrained multi-view reconstruction model, VGGT~\citep{wang2025vggt}, which provides both a 3D point cloud and the corresponding camera poses. 
Based on this geometry, we discretize the 3D space into a regular grid and define the set of active voxels $\{\mathbf{p}_i\}_{i=1}^L$, where each $\mathbf{p}_i \in \{0,1,\dots,n-1\}^3$ corresponds to the spatial index of a voxel in a discretized grid of resolution $n$ along each axis, and is occupied by the reconstructed point cloud.

To bridge 2D visual observations with spatially structured 3D representations, we lift image features into voxel space via multi-view aggregation. For each input view, we extract image features using a pretrained DINOv2~\citep{oquab2023dinov2} model and lift them into 3D space using the estimated camera parameters. For each active voxel center, we project its 3D coordinate onto every visible input view using the estimated camera parameters and sample the corresponding DINOv2 feature. 
These features are accumulated at voxel locations across views by average pooling, resulting in a per-voxel feature $\mathbf{f}_i$ that aggregates multi-view visual evidence. 
This feature lifting process establishes a direct correspondence between image observations and spatially localized 3D features, enabling appearance-aware modeling over the reconstructed geometry.

The resulting set of voxel features $\{(\mathbf{f}_i, \mathbf{p}_i)\}_{i=1}^L$ forms a unified representation of the object, where voxel positions provide coarse geometric structure and the associated features encode fine-grained visual details.

To further incorporate global context, we map the voxel features into a set of latent embeddings:
\begin{equation}
\mathbf{z} = \mathcal{E}(\{(\mathbf{f}_i, \mathbf{p}_i)\}_{i=1}^L), \quad \mathbf{z} = \{(\mathbf{z}_i, \mathbf{p}_i)\}_{i=1}^L,
\end{equation}
where $\mathbf{z}_i \in \mathbb{R}^d$ denotes the embedding associated with voxel $\mathbf{p}_i$. 
The encoder $\mathcal{E}$ is implemented as a transformer operating on the voxel tokens, enabling context-aware interaction across the 3D structure.

This voxel-based feature representation, together with the feature lifting mechanism, serves as the foundation for jointly predicting geometry and spatially varying physical properties in the following stage.

\subsubsection{Joint Geometry and Physics Prediction}

Given the voxel-based latent representation $\mathbf{z} = \{(\mathbf{z}_i, \mathbf{p}_i)\}_{i=1}^L$, we jointly decode geometry, appearance, and physical properties.

\emph{1) Geometry and appearance decoding.}
We first map the latent representation to an explicit 3D representation for rendering. 
We adopt a 3D Gaussian representation, where each latent embedding $\mathbf{z}_i$ is decoded into a set of Gaussian primitives:
\begin{equation}\label{eq:3dgs_decoder}
\mathcal{G} = \mathcal{D}_{\text{geo}}(\mathbf{z}),
\end{equation}
where $\mathcal{D}_{\text{geo}}$ predicts the 3D Gaussian parameters~\citep{kerbl20233d}, including position, scale, opacity, and appearance attributes. 
This formulation enables efficient rendering while preserving fine-grained geometric and visual details.

\emph{2) Physical property prediction.} 
In parallel, we interpret the latent representation $\mathbf{z}$ as a spatial field and predict physical attributes at each voxel location. 
Specifically, the physical properties associated with each voxel are predicted as:
\begin{equation}\label{eq:phys_decoder}
\mathbf{P} = \mathcal{D}_{\text{phys}}(\mathbf{z}, \mathbf{c}_{\text{phys}}),
\end{equation}
where $\mathbf{P} = \{\mathbf{P}_i\}_{i=1}^{L}$ denotes voxel-wise physical properties. Each element is defined as
\begin{equation}
\mathbf{P}_i = [\mathbf{E}_i, \boldsymbol{\nu}_i, \boldsymbol{\rho}_i, \mathbf{M}_i],
\end{equation}
including Young’s modulus $\mathbf{E}_i$, Poisson’s ratio $\boldsymbol{\nu}_i$, density $\boldsymbol{\rho}_i$, and material category $\mathbf{M}_i$. 
The optional condition $\mathbf{c}_{\mathrm{phys}}$ encodes a high-level material description, such as ``soft'' or ``rigid'' and is incorporated through cross-attention to modulate voxel features. This semantic conditioning provides additional control over the predicted physical properties without requiring users to manually specify individual numerical parameters.

Our joint prediction framework defines a spatially varying physical field over the object, where local physical properties are inferred from spatially grounded features.
Since both the geometry $\mathcal{G}$ and the physical properties $\mathbf{P}$ are derived from the same latent representation $\mathbf{z}$,
the resulting geometry and material attributes are intrinsically aligned, enabling consistent and simulation-ready 3D asset generation.

\subsubsection{Single-View Feature Generation}
The above method relies on multi-view images, which may not be available in many practical scenarios. To enable single-view reconstruction as in PhysGM~\citep{lv2025physgm}, we learn to directly generate the 3D latent representation $\mathbf{z}$ from a single image using a conditional flow matching formulation~\citep{lipman2022flow}.

Given a training set of 3D objects, we first render multi-view training images $\{\mathbf{I}_i\}_{i=1}^N$ and construct the corresponding 3D representation $\mathbf{z}$ using the multi-view pipeline described in Sec.~\ref{Voxel-Aligned Feature Construction}. 
We then randomly select a single view $\mathbf{I}$ as the conditioning input and train a conditional velocity field $\mathbf{u}_\theta$ that models the evolution from Gaussian noise to $\mathbf{z}$.
Specifically, we sample a Gaussian noise $\mathbf{x}_0$ and define a linear interpolation path:
\begin{equation}
\mathbf{x}_t = (1 - t)\mathbf{x}_0 + t\mathbf{z}, \quad t \in [0,1].
\end{equation}
The target velocity along this path is given by $\mathbf{u}^* = \mathbf{z} - \mathbf{x}_0$.
The model $\mathbf{u}_\theta(\mathbf{x}_t, t, \mathbf{I}_k)$ is trained to predict this velocity $\mathbf{u}^*$ conditioned on the image $\mathbf{I}_k$~\citep{lipman2022flow}.

At test time, given only a single image $\mathbf{I}$, we generate the latent representation $\mathbf{z}$ by integrating the learned velocity field:
\begin{equation}
\frac{d\mathbf{x}_t}{dt} = \mathbf{u}_\theta(\mathbf{x}_t, t, \mathbf{I}), \quad \mathbf{x}_0 \sim \mathcal{N}(0, {I_d}),
\end{equation}
from $t=0$ to $t=1$, where $I_d$ denotes the identity variance matrix. The final state $\mathbf{x}_1$ is used as the generated 3D feature.

This formulation enables direct generation of 3D latent features from a single image, bypassing the need for multi-view inputs.
Importantly, the generated features lie in the same representation space as the multi-view reconstructed features. Consequently, the generated representation can be decoded using the same
geometry and physical property heads as the multi-view representation.

\begin{table*}[t]
\centering
\small
\setlength{\tabcolsep}{4.5pt}
\renewcommand{\arraystretch}{1.12}
\resizebox{\textwidth}{!}{%
\begin{tabular}{l cc cc cc cc cc cc c}
\toprule
& \multicolumn{2}{c}{\textbf{Flower}}
& \multicolumn{2}{c}{\textbf{Metal Crate}}
& \multicolumn{2}{c}{\textbf{Sand}}
& \multicolumn{2}{c}{\textbf{Rubber Toy}}
& \multicolumn{2}{c}{\textbf{Wooden Box}}
& \multicolumn{2}{c}{\textbf{Average}}
& \multirow{2}{*}{\makecell[c]{\textbf{Runtime Cost}}} \\
\cmidrule(lr){2-3}
\cmidrule(lr){4-5}
\cmidrule(lr){6-7}
\cmidrule(lr){8-9}
\cmidrule(lr){10-11}
\cmidrule(lr){12-13}
\textbf{Method}
& $\mathrm{CLIP}_{\mathrm{sim}}\uparrow$
& $\mathrm{U}_{\mathrm{score}}\uparrow$
& $\mathrm{CLIP}_{\mathrm{sim}}\uparrow$
& $\mathrm{U}_{\mathrm{score}}\uparrow$
& $\mathrm{CLIP}_{\mathrm{sim}}\uparrow$
& $\mathrm{U}_{\mathrm{score}}\uparrow$
& $\mathrm{CLIP}_{\mathrm{sim}}\uparrow$
& $\mathrm{U}_{\mathrm{score}}\uparrow$
& $\mathrm{CLIP}_{\mathrm{sim}}\uparrow$
& $\mathrm{U}_{\mathrm{score}}\uparrow$
& $\mathrm{CLIP}_{\mathrm{sim}}\uparrow$
& $\mathrm{U}_{\mathrm{score}}\uparrow$
\\
\midrule
DreamPhysics
& 26.92 & 2.77
& 25.03 & 2.35
& 20.86 & 2.85
& 27.39 & 3.08
& 20.86 & 1.62
& 24.212 & 2.534
& $>1$\,h \\

OmniPhysGS
& 25.50 & 1.46
& 25.21 & 1.88
& 22.29 & 1.88
& 26.19 & 2.38
& 20.38 & 2.08
& 23.914 & 1.936
& $>5$\,h \\

PhysGM
& 24.08 & 3.00
& 25.07 & \underline{3.31}
& 22.66 & 2.12
& 27.83 & 3.19
& 25.02 & 2.42
& 24.932 & 2.808
& $\sim10$\,s \\

\specialrule{0.3pt}{1pt}{1pt}
\textbf{PhysCreator}
& \textbf{28.09} & \underline{4.12}
& \underline{27.62} & \textbf{4.35}
& \textbf{25.05} & \textbf{3.69}
& \underline{28.08} & \underline{4.23}
& \underline{26.08} & \underline{4.27}
& \underline{26.984} & \underline{4.132}
& $\sim10$\,s \\

\addlinespace[2pt]
\textbf{PhysCreator$^*$}
& \underline{27.13} & \textbf{4.38}
& \textbf{27.96} & \textbf{4.35}
& \underline{23.17} & \underline{3.08}
& \textbf{28.65} & \textbf{4.58}
& \textbf{28.10} & \textbf{4.46}
& \textbf{27.002} & \textbf{4.170}
& $\sim15$\,s \\
\bottomrule
\end{tabular}%
}
\caption{Quantitative comparison across multiple object categories in terms of semantic consistency ($\mathrm{CLIP}_{\mathrm{sim}}, \%$), user-study score ($\mathrm{U}_{\mathrm{score}}$), and per-object runtime cost. {PhysCreator$^*$} denote our model with multi-view input. Best and second-best results are shown in bold and underlined, respectively. The proposed algorithm achieves consistently favorable semantic and human-evaluation scores while maintaining efficient inference.}
\label{tab:clip_user_study}
\end{table*}

\subsection{Vision-Language Distillation}
\label{Vision-Language Distillation}
Learning spatially varying physical properties requires fine-grained supervision, which is difficult to obtain at scale. 
While simulation-based training with SDS loss~\citep{huang2025dreamphysics,lin2025omniphysgs,zhang2024physdreamer} can provide physically grounded supervision, it is computationally expensive and often too weak to train a generalizable model from scratch~\citep{li2024instant3d}. 
In this work, we propose a strategy that uses vision-language distillation to extract physical knowledge from pretrained VLMs~\citep{comanici2025gemini} for large-scale physical training.

Given images rendered from 3D objects, we first use a VLM~\citep{comanici2025gemini} to analyze rendered images and infer a set of semantic parts together with associated physical attributes. 
This produces a set of semantic descriptions $\{\textsc{text}_c\}_{c=1}^C$ with the corresponding physical parameters $\{\mathbf{\tilde{P}}_c\}_{c=1}^C$ consisting of Young’s modulus, Poisson’s ratio, density, and material category.
To transfer these predictions to 3D, we leverage CLIP~\citep{radford2021learning} as a shared embedding space. 
We extract dense image features for each view and lift them into voxel space via projection and aggregation, yielding a feature vector $\mathbf{vox}_i$ for each voxel. 
In parallel, we encode each semantic description $\textsc{text}_c$ into a text embedding using the CLIP text encoder.

We then assign a semantic label to each voxel by computing cosine similarity between voxel features and text embeddings:
\begin{equation}
c^*(i) = \arg\max_c \; \langle \mathbf{vox}_i, \mathbf{tex}_c \rangle,
\end{equation}
where $\mathbf{tex}_c$ denotes the CLIP text embedding of $\textsc{text}_c$. 
The physical properties of voxel $i$ are then obtained by
\begin{equation}
\mathbf{\hat{P}}_i = [\mathbf{\hat{E}}_i, \boldsymbol{\hat{\nu}}_i, \boldsymbol{\hat{\rho}}_i, \mathbf{\hat{M}}_i] = \mathbf{\tilde{P}}_{c^*(i)}.
\end{equation}

This process yields voxel-wise supervision that encodes spatially varying physical attributes aligned with visual appearance. 
To improve the reliability of this supervision, we further perform an offline simulation-based verification step. Specifically, we run physical simulations using the inferred properties $\mathbf{\hat{P}}$ and ask the VLM to assess whether the resulting dynamic behaviors are consistent with the corresponding material semantics. 
Only samples that pass this verification are retained for reliable training.

We train our model by distilling knowledge from $\mathbf{\hat{P}}$: 
\begin{equation}
\mathcal{L}_{\text{distill}}
= \sum_{i=1}^{L} \Big(
 \|\mathbf{E}_i - \hat{\mathbf{E}}_i\|_2^2
+ \|\boldsymbol{\nu}_i - \hat{\boldsymbol{\nu}}_i\|_2^2 
 + \|\boldsymbol{\rho}_i - \hat{\boldsymbol{\rho}}_i\|_2^2
+ \mathcal{L}_{\text{CE}}
  (\mathbf{M}_i, \hat{\mathbf{M}}_i)
\Big),
\label{eq:distill_loss}
\end{equation}
where continuous physical parameters are optimized using mean squared error, and discrete material categories are trained with a cross-entropy loss $\mathcal{L}_{\text{CE}}$.
For numerical stability, Young’s modulus $\mathbf{E}$ and density $\boldsymbol{\rho}$ are transformed to log scale and normalized.

In addition, the inferred semantic and physical attributes can be further expressed as textual descriptions, which are used as the optional conditioning signal $\mathbf{c}_{\text{phys}}$ in Eq.~\ref{eq:phys_decoder} for controllable manipulation of material properties.

This distillation framework enables scalable training by transferring high-level physical reasoning from VLMs into spatially grounded 3D fields, which allows PhysCreator to learn generalizable mappings from visual appearance to spatially varying material properties.

Overall, PhysCreator jointly predicts 3D Gaussians and spatially varying physical fields through a shared voxel-aligned representation, producing simulation-ready assets via feed-forward inference.

\section{Experiments}
\subsection{Experimental settings}

\paragraph{Simulation Details.}
We adopt the MPM~\citep{jiang2015affine, jiang2016material} implementation from PhysGaussian~\citep{xie2024physgaussian} as the physics simulator. The simulator takes as input a Gaussian point cloud augmented with physical properties, where each particle is associated with a material type and continuous parameters including Young’s modulus, Poisson’s ratio, and density. External forces such as gravity and wind are applied to produce dynamic behaviors.
Similar to PhysGM~\citep{lv2025physgm}, we use Objaverse~\citep{deitke2023objaverse} for training and evaluation. 
For each object, we extract mesh assets and convert them into the input format required by each method.

\paragraph{Baselines.}
We compare PhysCreator against the optimization-based methods DreamPhysics~\citep{huang2025dreamphysics} and OmniPhysGS~\citep{lin2025omniphysgs}, as well as the feed-forward method PhysGM~\citep{lv2025physgm}. We use the official implementations and recommended configurations for all baseline methods. DreamPhysics and OmniPhysGS require per-instance test-time optimization, meaning their model has to be retrained for each new input.
In contrast, PhysGM and PhysCreator requires only one forward pass, without any input-specific optimization or parameter updates at test time.

\begin{figure}[t]
  \centering
  \includegraphics[width=.95\columnwidth]{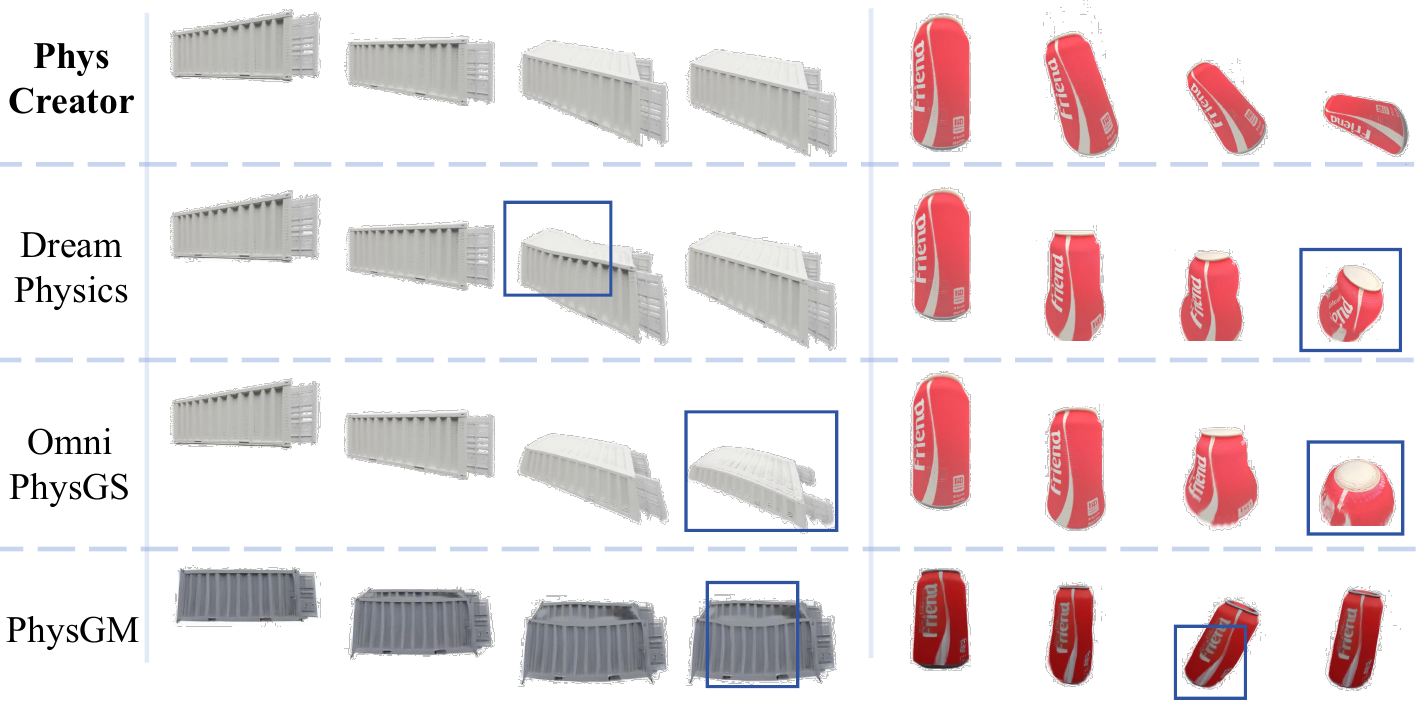}
  \caption{
  Qualitative comparison of simulated dynamics under gravity.
  PhysCreator produces stable and physically consistent motion with realistic contact and deformation. In contrast, OmniPhysGS exhibits structural collapse after impact with the ground, while DreamPhysics and PhysGM show unrealistic local deformation.
  }
  \label{fig:comparison1}
\end{figure}

\begin{figure}[t]
  \centering
\includegraphics[width=\columnwidth]{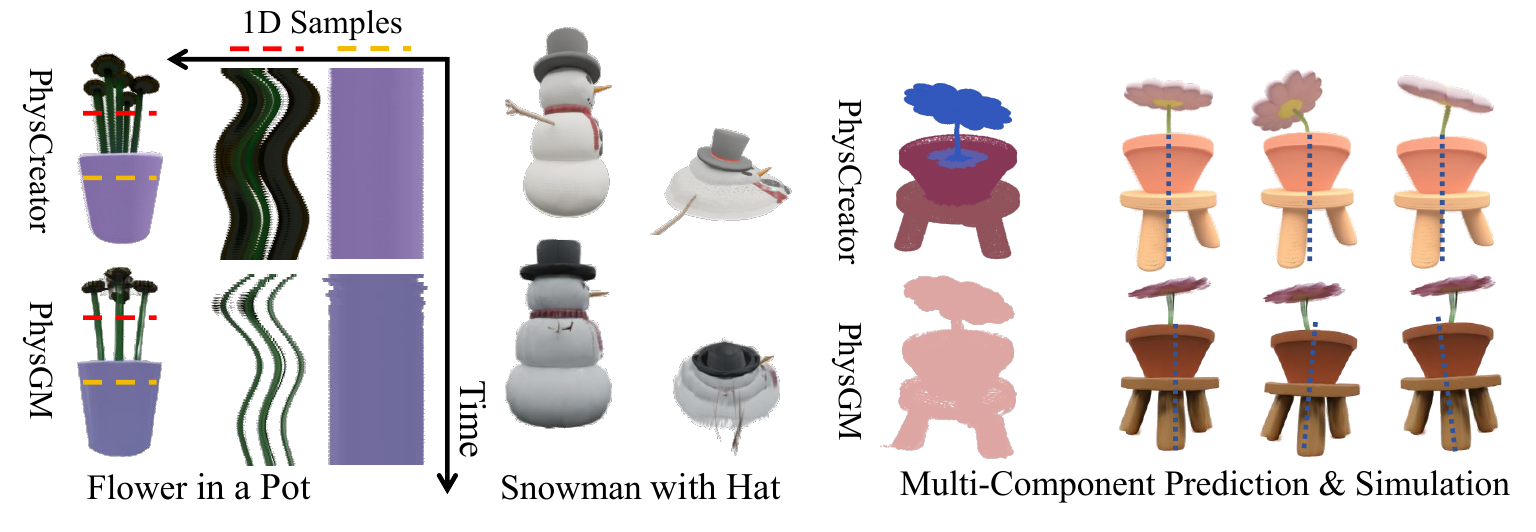}
\caption{ 
Effectiveness of spatially varying physical property modeling. 
\textbf{Top}: PhysCreator enables the flower to sway while keeping the pot stable, and allows the snow body to deform while preserving the hat; PhysGM produces implausible deformation due to uniform properties. 
1D samples across time (space--time slices highlighted by dotted lines) are included to better visualize the temporal dynamics.
\textbf{Bottom}: PhysCreator localizes motion to the flower while keeping the pot and chair stable, whereas PhysGM causes all components to move together. The blue dashed axes highlight this unintended coupled motion.
}
\label{fig:vs_physGM}
\end{figure}

\subsection{Results}

Following the evaluation protocol of PhysGM~\citep{lv2025physgm}, we use CLIP-based semantic similarity ($\text{CLIP}_{\text{sim}}$) and user-study scores ($\text{U}_{\text{score}}$) as quantitative metrics. $\text{CLIP}_{\text{sim}}$ measures the semantic alignment between the rendered simulation outcomes and their corresponding textual descriptions of the physical phenomena, while $\text{U}_{\text{score}}$ reflects human judgments of perceptual quality and physical realism. Our user study involves 26 participants and is conducted under a blind setting using a 5-point Likert scale.
We evaluate all baselines using their official implementations and recommended configurations. 

As shown in Table~\ref{tab:clip_user_study}, PhysCreator compares favorably with state-of-the-art approaches~\citep{huang2025dreamphysics,lin2025omniphysgs,lv2025physgm}. 
PhysCreator with multi-view input achieves the highest average scores, benefiting from the additional visual information provided by multiple views. Nevertheless, PhysCreator with a single view closely matches its multi-view counterpart (PhysCreator$^*$), demonstrating that the single-view generative model preserves high quality despite receiving substantially less input information.

Moreover, PhysCreator consistently outperforms PhysGM, which demonstrates that PhysCreator more effectively recovers semantically appropriate and perceptually plausible physical behaviors from a single image.

Qualitative comparisons in Fig.~\ref{fig:comparison1} further show that PhysCreator produces more stable and physically consistent dynamics, whereas baseline methods exhibit artifacts such as structural collapse, excessive bouncing, or implausible local deformation under external forces. The observed failure modes are qualitatively consistent with those reported in prior comparative evaluations~\citep{lv2025physgm}.

In terms of runtime, DreamPhysics and OmniPhysGS optimize a separate model for each test object, incurring several hours of computation per instance. In contrast, PhysGM and PhysCreator directly generalize to unseen objects with a single shared model. Without object-specific optimization or parameter updates, PhysCreator produces a simulation-ready 3D asset in seconds through a single feed-forward pass.
Together, these results demonstrate that PhysCreator achieves a favorable balance between physical plausibility and computational efficiency. In particular, its single-view model approaches the performance of multi-view reconstruction while substantially outperforming PhysGM.

\subsection{Analysis and Discussion}

\subsubsection{Spatially Varying Material Modeling}

Modeling spatially varying material properties is crucial for achieving realistic dynamics in composite objects. 
PhysCreator predicts voxel-level physical attributes, enabling different object parts to respond independently according to their material characteristics. 
As illustrated in Fig.~\ref{fig:vs_physGM}, the flower exhibits natural swaying while the pot remains stable under wind forces, and the snow deforms realistically while the hat preserves its structure. 
By comparison, methods that assume globally uniform materials, such as PhysGM, often produce unrealistic deformation or structural collapse in similar scenarios. 
This part-aware material modeling produces flexible yet stable motion across all components, resulting in physically plausible and visually coherent simulations. 
These results highlight the effectiveness of spatially varying material modeling for generating simulation-ready 3D assets.

\subsubsection{Geometry and Physical-Property Learning}
We first evaluate the quality of the reconstructed geometry. 
Chamfer Distance measures the discrepancy between the reconstructed and ground-truth surfaces, while F-score at a $2\%$ distance threshold evaluates surface accuracy and completeness.
As shown in Table~\ref{tab:reconstruction_quality}, PhysCreator reduces the Chamfer Distance and improves the F-score. These results demonstrate that the proposed voxel-aligned representation supports accurate geometry learning.

We further evaluate the accuracy of the predicted physical properties using a set of manually verified reference annotations. The initial property values are inferred by a pretrained VLM and subsequently inspected through rendering and physical simulation. Human evaluators then assess whether the assigned properties produce physically plausible dynamic behaviors.
We adopt a conservative acceptance criterion and retain only annotations that are consistently judged to be plausible.
As shown in Table~\ref{tab:physical_property_error}, PhysCreator achieves the lowest prediction errors for the evaluated physical properties, showing the effectiveness of the proposed algorithm.

\begin{table}[t]
\centering
\small
\setlength{\tabcolsep}{5pt}
\renewcommand{\arraystretch}{1.12}
\begin{tabular}{lcc}
\toprule
Method
& Chamfer Distance$\downarrow$
& F-score @ 2\%$\uparrow$ \\
\midrule

PhysGM~\citep{lv2025physgm}
& 0.0475
& 59.05 \\

PhysCreator
& \textbf{0.0431}
& \textbf{72.72} \\

\bottomrule
\end{tabular}

\caption{
Quantitative comparison of geometry reconstruction quality.
PhysCreator achieves a lower Chamfer Distance and a higher F-score,
indicating more accurate and complete surface reconstruction.
}
\label{tab:reconstruction_quality}
\end{table}

\begin{table}[t]
\centering
\small
\setlength{\tabcolsep}{4.5pt}
\renewcommand{\arraystretch}{1.12}
\begin{tabular}{l ccc}
\toprule
Physical Property
& {DreamPhysics}
& {PhysGM}
& {PhysCreator} \\
\midrule
Young's modulus
& 13.40\%
& 16.19\%
& \textbf{7.78\%} \\

Density
& --
& --
& \textbf{11.86\%} \\

Poisson ratio
& --
& 18.66\%
& \textbf{8.12\%} \\
\bottomrule
\end{tabular}%
\caption{
Mean relative errors ($\downarrow$) of physical-property predictions.
Young's modulus and density are evaluated in logarithmic space because of their large dynamic ranges, whereas Poisson's ratio is evaluated in its original scale.
``--'' indicates that the corresponding method does not predict this property, using a manually selected value for simulation instead. 
}
\label{tab:physical_property_error}
\end{table}

\subsubsection{Controllable Physical Generation}

\begin{table}[!t]
\centering
\small
\setlength{\tabcolsep}{6pt}
\renewcommand{\arraystretch}{1.15}
\begin{tabular}{ccc}
\toprule
\makecell[c]{Property Ratio} & \makecell[c]{PhysCreator} & \makecell[c]{DreamPhysics} \\
\midrule
$E_{\mathrm{hard}}/E_{\mathrm{soft}}$ & \textbf{13.18} & 1.70 \\
$\rho_{\mathrm{heavy}}/\rho_{\mathrm{light}}$ & \textbf{11.60} & 1.00 \\
\bottomrule
\end{tabular}
\caption{Quantitative comparison of text-conditioned physical control. Higher property ratios indicate stronger responsiveness to the corresponding semantic prompts. PhysCreator demonstrates substantially stronger control than the baseline method~\citep{huang2025dreamphysics}.}
\label{tab:text_control}
\end{table}

\begin{figure}[!t]
  \centering
\includegraphics[width=\columnwidth]{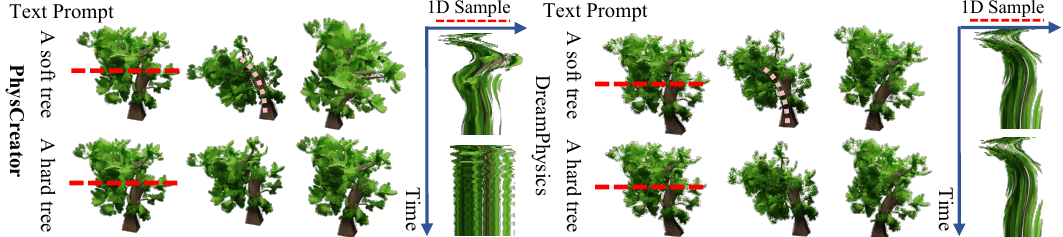}
\caption{Qualitative comparison of text-conditioned physical control under identical wind forces. PhysCreator produces distinct prompt-consistent bending for ``soft'' and ``hard,'' whereas DreamPhysics shows limited responsiveness. The pink dashed curves indicate bending trajectories.}
\label{fig:hard_soft}
\end{figure}

We evaluate the effectiveness of text conditioning for controlling physical properties. 
As shown in Table~\ref{tab:text_control}, PhysCreator exhibits substantially higher sensitivity to textual prompts compared to DreamPhysics~\citep{huang2025dreamphysics}, which also supports text-based control. 
This result indicates that PhysCreator can effectively adjust physical parameters in accordance with high-level semantic descriptions, while the baseline shows limited responsiveness. 

Fig.~\ref{fig:hard_soft} provides qualitative comparisons under different prompts.  
The figure shows simulated dynamics of a tree under wind forces with different textual prompts. 
With text conditioning, our model produces distinct and physically consistent behaviors aligned with the prompts, yielding stiffer motion under ``hard'' and more flexible deformation under ``soft''. 
In contrast, DreamPhysics results in nearly indistinguishable dynamics across different prompts, indicating a lack of effective control.
These results demonstrate that the proposed conditioning mechanism enables controllable physical generation from high-level semantic descriptions without manual tuning of individual physical parameters.

\section{Conclusion}

We present PhysCreator, a unified framework for generating simulation-ready 3D assets with both geometry and spatially varying physical properties from images. By lifting visual observations into a voxel-aligned latent representation, PhysCreator jointly models geometry, appearance, and physical attributes within a shared 3D space. To alleviate scarce physical annotations, we distill material-aware VLM knowledge into voxel-level supervision and improve pseudo-label reliability through simulation-based verification. The learned representation further supports semantic control over coupled physical attributes. Experiments show that PhysCreator efficiently generates high-quality, physically plausible, and simulation-ready 3D assets across diverse objects.

\newpage
\clearpage

\bibliographystyle{iclr2025_conference}
\bibliography{ref}

\end{document}